\documentclass[letterpaper, 10 pt, conference]{ieeeconf}  

\IEEEoverridecommandlockouts                              

\usepackage{graphbox,graphicx}
\usepackage{tikz}
\usepackage{bm}
\usepackage{pgfplots}
\usepackage{amsmath}
\usepackage{mathtools}
\pgfplotsset{compat=1.17}
\usepackage{nicefrac}
\usepackage[subrefformat=parens]{subcaption}
\usepackage[noadjust]{cite}

\title{\LARGE \bf Pixel State Value Network for Combined Prediction and Planning in Interactive Environments}

\author{Sascha Rosbach$^{1,2}$, Stefan M. Leupold$^{1}$, Simon Gro{\ss}johann$^{1}$ and Stefan Roth$^{2}$ %
\thanks{$^{1}$The authors are with CARIAD SE, 38440
        Wolfsburg, Germany
        {\tt\small  \{sascha.rosbach, stefan.leupold, simon.grossjohann\}@cariad.technology}}%
\thanks{$^{2}$The authors are with the Visual Inference Lab,
        Department of Computer Science, Technische Universit\"at Darmstadt,
        64289 Darmstadt, Germany
        {\tt\small stefan.roth@visinf.tu-darmstadt.de}}%
}

\begin{document}

\maketitle
\thispagestyle{empty}
\pagestyle{empty}


\begin{abstract}
    Automated vehicles operating in urban environments have to reliably interact with other traffic participants.
    Planning algorithms often utilize separate prediction modules forecasting probabilistic, multi-modal, and interactive behaviors of objects.
    Designing prediction and planning as two separate modules introduces significant challenges, particularly due to the interdependence of these modules.
    This work proposes a deep learning methodology to combine prediction and planning. 
    A conditional GAN with the U-Net architecture is trained to predict two high-resolution image sequences.
    The sequences represent explicit motion predictions, mainly used to train context understanding, and pixel state values suitable for planning encoding kinematic reachability, object dynamics, safety, and driving comfort.
    The model can be trained offline on target images rendered by a sampling-based model-predictive planner, leveraging real-world driving data.
    Our results demonstrate intuitive behavior in complex situations, such as lane changes amidst conflicting objectives.
\end{abstract}


\section{Introduction}
	
    State-of-the-art automated driving functions utilize a sequential processing chain consisting of perception, prediction, planning, and control.
    In this chain, the interdependence of prediction and planning presents a major challenge.
    It is often neglected that the actions of the automated vehicle influence the behavior of other traffic participants.
    This work concentrates on the joint prediction of both object behavior (other agents) and ego behavior (the automated vehicle).
    
\begin{figure}  
  \vspace{1.8mm}
  \centering
  \includegraphics[width=0.48\textwidth]{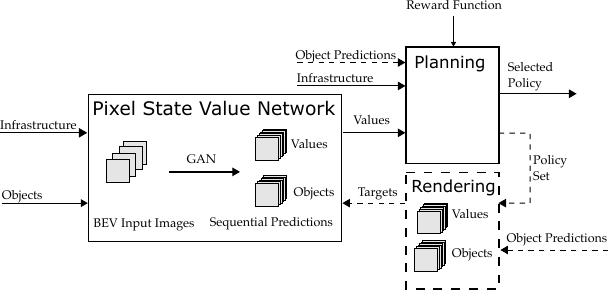}
  \caption{ Shows the proposed architecture comprised of pixel state value network (PSVN), planning, and rendering module.
            The rendering module uses policy sets and object predictions to generate targets for offline training of the PSVN. 
            The PSVN infers pixel state values for the planning module.
            Object predictions and the rendering module (dashed lines) are not required during inference.
  }
  \label{fig:title}
\end{figure}

    Due to inherent uncertainty in real-world driving situations, it would be required to provide a planner with probabilistic multi-modal predictions that jointly provide a coherent future.
    However, implementing the desired driving style based on uncertain motion predictions is difficult and an error-prone task.
    In contrast to related work in motion prediction, we do not predict trajectories but, instead, focus on predicting two time-dependent image sequences implicitly encoding joint object and ego behavior.
    High-resolution image sequences allow us to represent arbitrary distributions over a pixel state space. 
    The first sequence contains future pixel visitations of other agents. 
    We use it as a training target to force the network to form a coherent world view but we do not utilize it for planning.
    The second sequence encodes the value of a position for the automated vehicle in space and time and can be directly used by a planner.
    The deep learning task is posed as paired image-to-image prediction task~\cite{isolaImagetoImageTranslationConditional2018} and uses four bird's eye view images as input that depict information about road infrastructure and the current state of objects and ego vehicle in the environment.
    The architecture is depicted in Fig.~\ref{fig:title}.
    
    In this work, we propose a novel imitation learning methodology that is able to scale in terms of experience and resolution of the state evaluation.
    The approach can be divided into two stages:
    The first stage is concerned with rendering image pairs and the second stage focuses on training the network offline.
    In the first stage, we create our dataset by replaying 24 hours of real-world driving data and use our 'exhaustive' model-predictive planner to render megapixel-sized value images using inputs from perception, localization, and predictions from queried object motion.
    In the second stage, we train a conditional GAN offline on the dataset (approx.\ 700~k images).
     
    We perform a qualitative evaluation of the image-to-image translation in challenging scenarios.
    Further, we perform a quantitative evaluation of collision avoidance and desired trajectory selection by contrasting the implicit object handling via predicted pixel state values against a traditional processing chain relying on explicit object predictions.
    

\section{Related Work}

    Traditionally, planning algorithms rely on prediction modules providing the future motion of traffic participants in the form of trajectories.
    State-of-the art motion forecasting approaches perform multi-modal predictions of agents using deep learning yielding a set of trajectories~\cite{luo2023jfp,varadarajan2022multipath++,shi2022mtr,gilles2021thomas}.
    A number of public benchmark datasets~\cite{chang2019argoverse,wilson2023argoverse,ettinger2021large,zhan2019interaction} have been proposed, setting the primary focus on the task of trajectory prediction with performance measured against ground truth.
    Planning algorithms use predictions to check for collisions~\cite{lavalle2006planning} and derive features such as proximities and time gaps over the planning horizon~\cite{mcnaughton2011parallel}.
    These features are used in the reward function to minimize risk and optimize the desired behavior.
    However, deriving such features based on uncertain behavior predictions is difficult. 
    In this work, we train a deep model that directly provides state values for planning in dynamic environments to produce the desired driving behavior.
    This circumvents the problem of generating interactive explicit object predictions by learning through observation of these features computed on recorded behaviors.
    
    Inverse Reinforcement Learning (IRL) is a methodology that concentrates on finding the unknown reward function for planning and Reinforcement Learning (RL) agents.
    Maximum Entropy Deep IRL (MEDIRL) predicts a reward map encoding the spatial traversability for path planning~\cite{wulfmeierLargescaleCostFunction2017}.
    Lee et al. propose an extension to MEDIRL for trajectory planning by predicting multiple time-dependent maps~\cite{lee2022spatiotemporal}.
    This allows to anticipate the behavior of moving objects implicitly without manually specifying a cost-function representation.
    However, reward maps generated by MEDIRL suffer from noise and artifacts due to utilized pixel state visitation frequency matching between demonstrations and model.
    This fundamental problem gets more severe when demonstrations are distributed on multiple target maps.
    Lee et al. propose a penalty loss for unvisited pixels to overcome these shortcomings~\cite{lee2022spatiotemporal}.
    However, the dimensionality of the observable space is limited, because the higher the spatio-temporal resolution, the less feedback received from the expert demonstrations.
    Wulfmeier et al. suggest first using human priors such as handcrafted reward maps to pretrain the model and then fine-tune using MEDIRL to address this issue~\cite{wulfmeier2016incorporating}.
    Our work addresses these problems and proposes a new methodology to discriminate states in a high-resolution spatio-temporal pixel space.
    Furthermore, we address the shortcoming of IRL being time-consuming to train because learning a representation for the reward function requires planning in the inner loop of reward learning.
    This gives our approach characteristics of value equivalence learning in Model-based Reinforcement Learning~\cite{grimm2020value,grimm2021proper, oh2017value,schrittwieser2020mastering}.
    However, our aim is to utilize learned state values to perform reward function shaping and later tune the influence using path integral IRL~\cite{rosbachDrivingStyleInverse2019}.
    
    A stream of literature aims to learn a holistic end-to-end model for driving by predicting a spatio-temporal cost volume for planning~\cite{zeng2019end,zeng2020dsdnet}.
    The authors train the approach using real-world driving data and define a max-margin planning loss function using the ground-truth ego trajectory as positive example and 100 randomly sampled trajectories as negative examples~\cite{zeng2020dsdnet}.
    This allows them to discriminate good from bad trajectories.
    We do not utilize demonstrations of the ego vehicle but, instead, 'exhaustively' sample kinematically feasible actions leading to trajectories having different behaviors (approx.\ 14~k trajectories per situation) and utilize the cost function of our planner to discriminate trajectories and also give our model the possibility to learn about reachability and collision avoidance.

\section{Methodology}
\label{sec:methodology}

    In this work, we propose a deep learning methodology for the combined prediction of both object behavior (other agents) and ego behavior (the automated vehicle).
    The training is formulated as a paired image-to-image translation task based on bird's eye view images of the environment.
    We first explain the image generation process that uses an exhaustive model-predictive planner at its core which is able to generate megapixel-sized target images.
    Furthermore, we propose a conditional GAN~\cite{isolaImagetoImageTranslationConditional2018} using the U-Net architecture~\cite{ronneberger2015u} to translate the image inputs to a sequence of images covering space and time.
    Last, we outline a hybrid mode of operation that combines value inference and sample-based planning.
    
\subsection{Image Pair Generation}
\begin{figure}[t]
     \vspace{1.8mm}
        \begin{subfigure}[b]{0.24\textwidth}
             \includegraphics[angle=9,width=\textwidth]{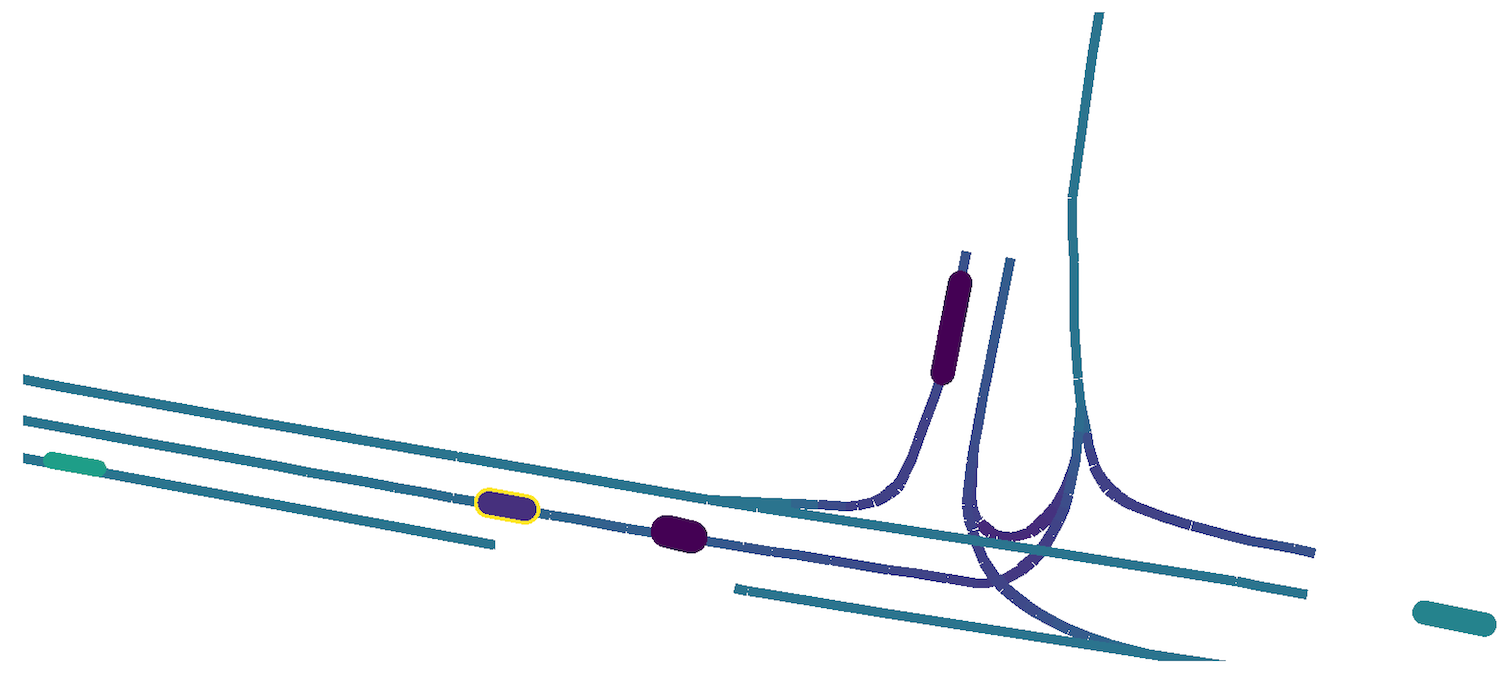}
            \caption{Velocity}
        \end{subfigure}
        \begin{subfigure}[b]{0.24\textwidth}
            \includegraphics[angle=9,width=\textwidth]{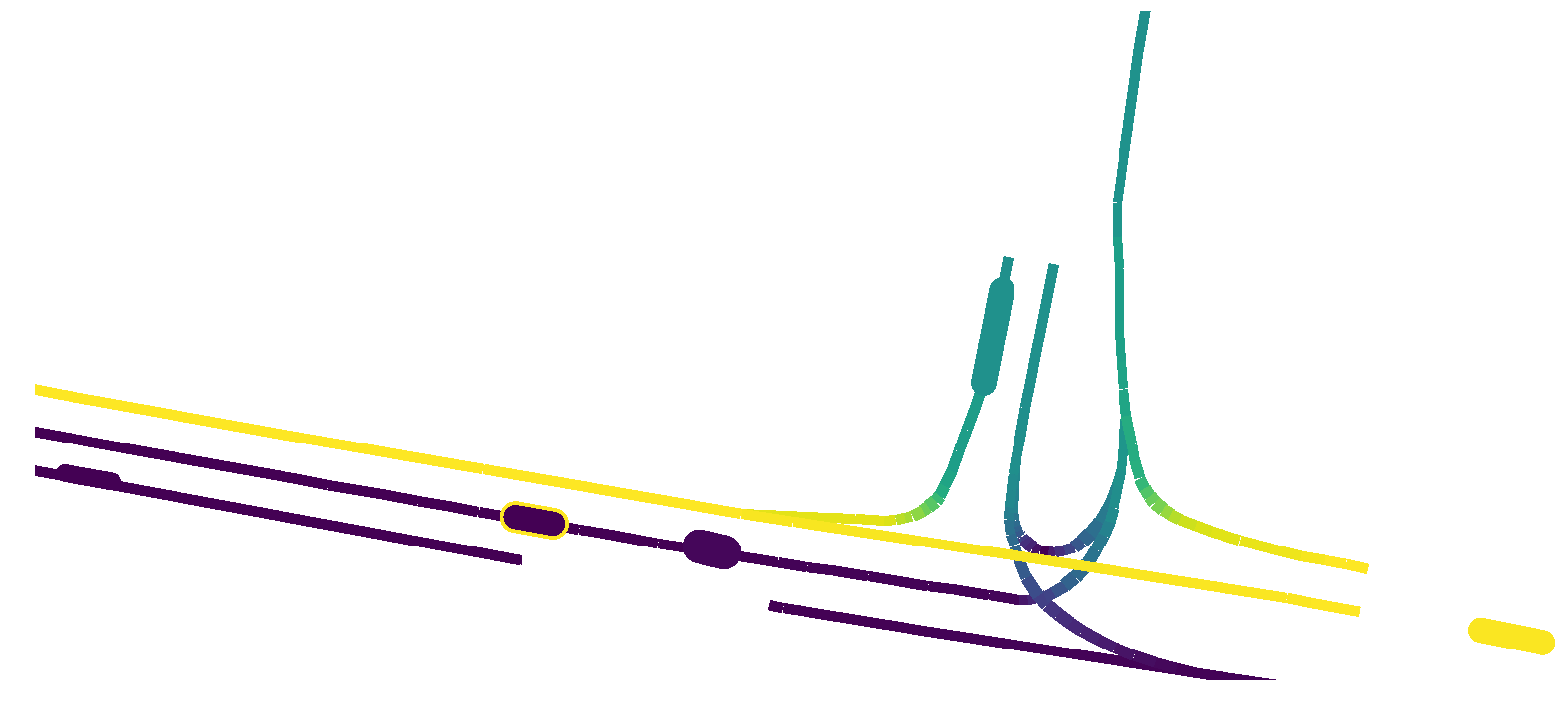}
            \caption{Direction}
        \end{subfigure}
        \begin{subfigure}[b]{0.24\textwidth}
            \includegraphics[angle=9,width=\textwidth]{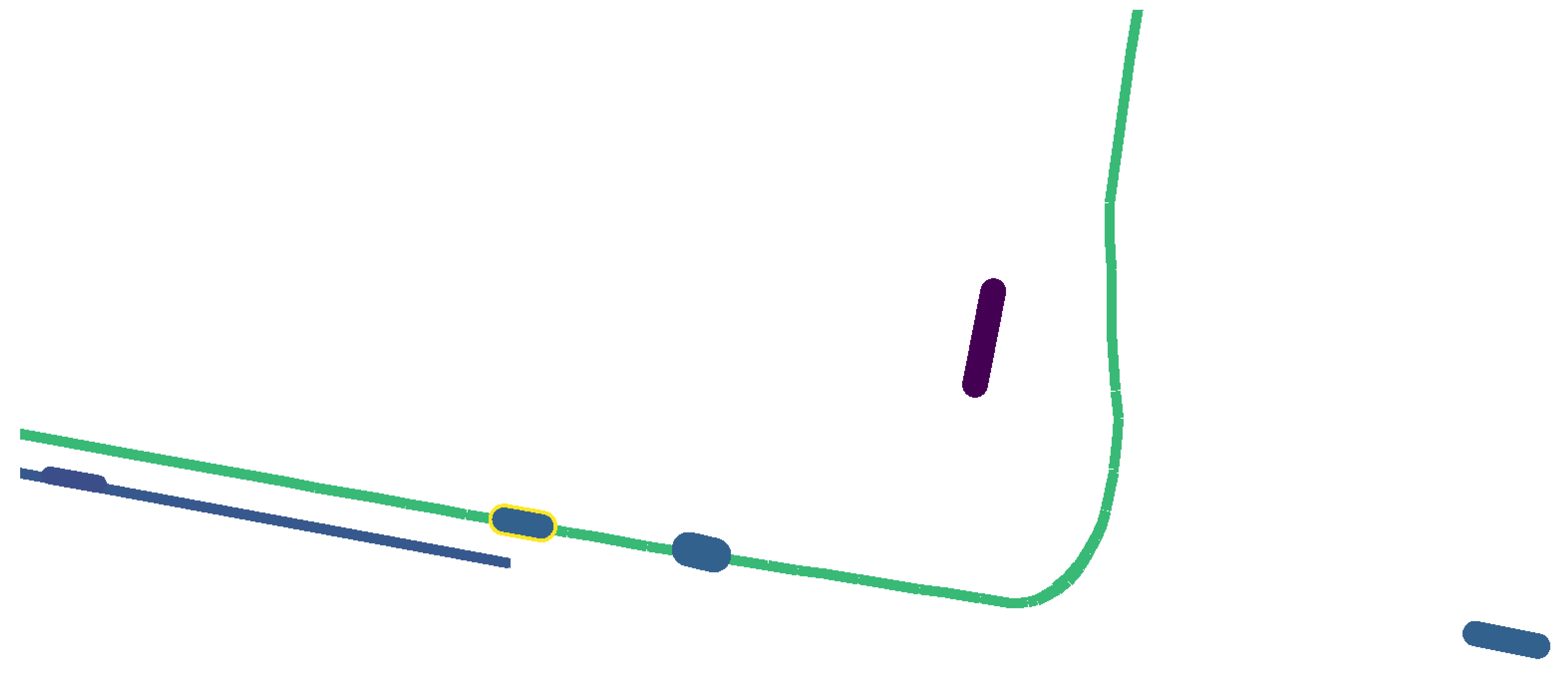}
            \caption{Target Lane}
        \end{subfigure}
        \begin{subfigure}[b]{0.24\textwidth}
            \includegraphics[angle=9,width=\textwidth]{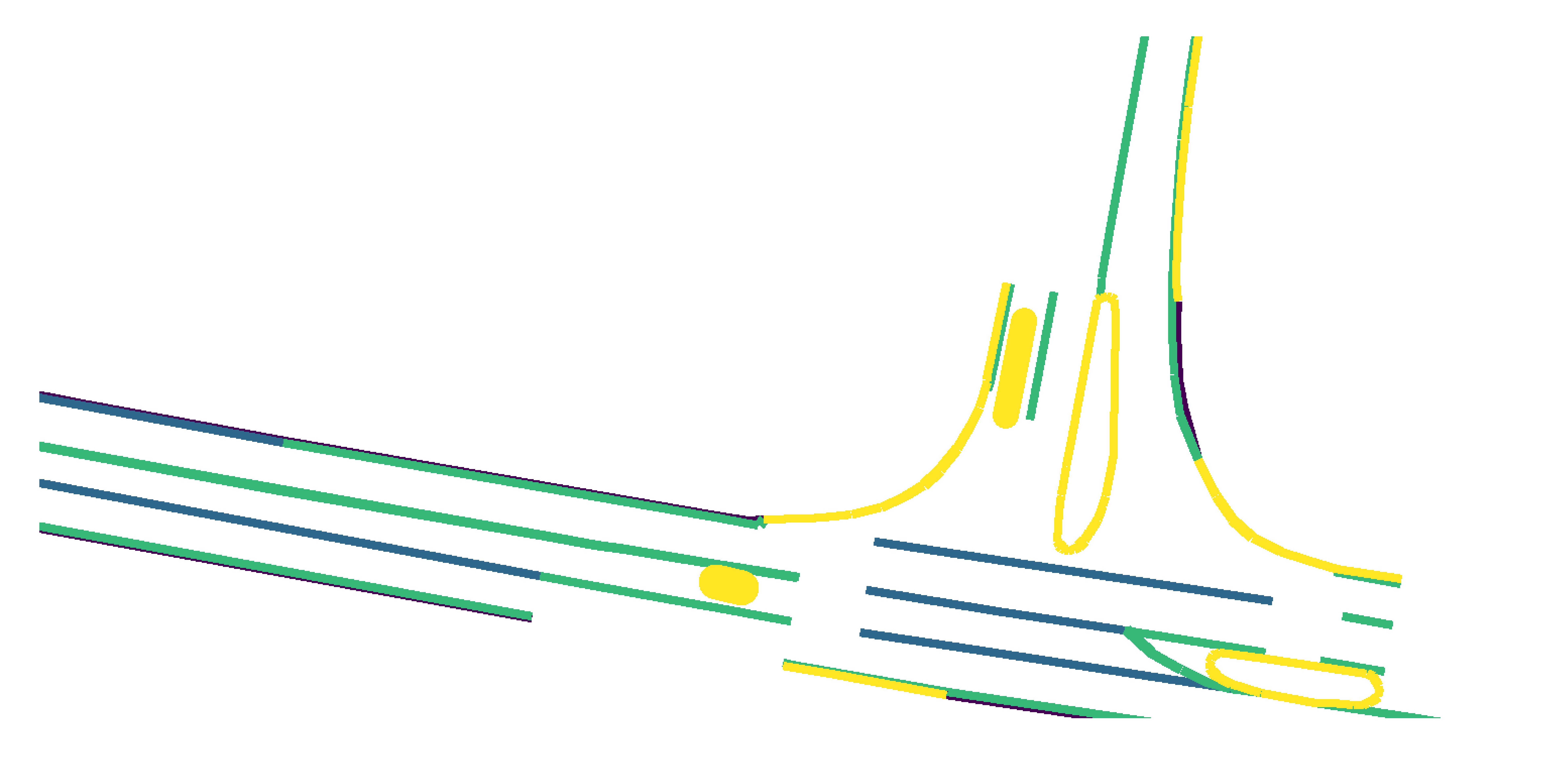}
            \caption{Boundaries}
        \end{subfigure}
        \caption{Depictions of the input representation for the neural network:
        \emph{(a)} Velocities of objects and speed limits of centerlines.
        \emph{(b)} Directions of objects and centerlines.
        \emph{(c)} Accelerations of objects and target lane.
        \emph{(d)} Static objects and boundaries.
        }
        \label{fig:four-input-images}
    \end{figure}

    Our approach allows using real-world data recordings.
    As of now, we chose to insert a layer of abstraction from raw sensory data.
    We assume access to a perception, a localization, and a road graph module to create bird's eye view input images.
    All images are rendered with one-megapixel resolution.
    The input consists of four images with a single channel each, stacked together as can be seen in Fig.~\ref{fig:four-input-images}.
    The first layer contains road infrastructure and velocity information for all objects including the ego vehicle.
    Road infrastructure is drawn with 10 pixel wide lines depicting the centerlines.
    Its color values correspond to speed limits and precomputed maximal velocities depending on road curvature and acceptable lateral acceleration.
    The ego vehicle is highlighted by an outline in the first three layers.
    The second layer depicts the direction of centerlines and all objects relative to the ego vehicle's orientation.
    The third layer contains accelerations of all objects and centerlines that are reachable by the ego vehicle.
    Centerlines are enumerated by color and the navigational target lane is highlighted.
    The fourth layer contains static vehicles and obstacles such as road boundaries classified as, e.g., dashed, solid, or curb.
    The deep neural network receives square images of size $512\times 512$.
    We resize the images without distortion by always rotating onto the diagonal to maintain the highest possible resolution in a square image.
    
    Target image generation can be separated into two algorithms: planning and rendering.
    Planning is performed at approximately $5$~Hz frequency on inputs of real-world driving recordings.
    Based on reprocessed data using perception, localization, and tracking, the planning algorithm generates a set of policies $\Pi$ and their corresponding values $V^{\Pi}$. 
    These are used by the rendering algorithm to produce images that approximate a value function over pixel coordinates.

\subsubsection{Planning algorithm}
    The planning algorithm is based on massively parallel search and runs on a graphics processing unit (GPU)~\cite{mcnaughton2011parallel,heinrich2018planning}. 
    Its pseudo code is given in Algo.~1.
    Implementation details of the planning algorithm have been published in our previous work on path integral IRL~\cite{rosbachDrivingStyleInverse2019,rosbachDrivingStyleEncoder2020,rosbach2020planning}.
    The algorithm starts by sampling a discrete number of continuous actions $\mathcal{A}_s$ (combining wheel angles and accelerations) for all states $s \in \mathcal{S}_t$.
    For each action, a transition is set up that uses fourth-order time-continuous polynomial functions for velocity and wheel angle profiles.
    The sampling distribution for each state is conditioned on feasible vehicle dynamics and the coefficients of the polynomials are derived from actor-friendly continuity constraints.
    The planner can generate diverse driving behavior by combining transitions of one to two seconds over the planning horizon of six seconds and more.
    This allows the policy set $\Pi$ to includes multiple behaviors, e.g., lane following, lane changes, swerving, and emergency stops. 
    To address the curse of dimensionality, a pruning operation removes redundant states that are evaluated based on state similarity, kinematic, and infrastructure based features.
    Our work makes use of a reward function with $K=25$ handcrafted features that can be categorized into motion, infrastructure, and object related components~\cite{rosbachDrivingStyleInverse2019,rosbachDrivingStyleEncoder2020,rosbach2020planning}.
    The planning algorithm yields a policy set $\Pi$ and computes a value \mbox{$V^{\pi}=\int_t \gamma_t R(s_t,a_t)\,dt$} for each $\pi \in \Pi$.
    We increase the state action space during the target image generation phase compared to deployment.
    This allows generating a large set of approx.\ 14~k policies (with a planning horizon of $6.6$ sec. and vehicle model integration steps of $0.2$ sec. this provides each situation with approx.\ 450~k states) that can be utilized to render dense pixel state value images.
    
 \subsubsection{Rendering algorithm}
    The rendering algorithm generates a sequence of images by drawing each of the planner's policies.
    The color of the pixels is derived from the policy value $V^{\pi}$ and discriminates pixels on the interval of [0,1] in each situation.
    The red, yellow, green coloring in Fig.~\ref{fig:qual2} can be seen as an example.
    The images have a resolution of one megapixel and the same aspect ratio as the input images. 
    The scaling and aspect ratio are determined by optimizing a convex hull around all object positions at time zero, the ego position, and a safety corridor around the reachable set of the last planning cycle.
    This provides situation adjusted, high-resolution pixel state value readings during inference.
    
    A sequence of images has six temporal layers $l$ for a planning horizon of 6.6 seconds with intervals $T_l$ as follows: 
    \begin{gather}
        T_l = ~ ](l-1)\times 1.1, l\times 1.1] \label{eq:temporal-layers} ,~ l \in \{1,\ldots,6\} .
    \end{gather}
    Before rendering, we calculate the probability $P$ of selecting a policy $\pi$ using the Maximum Entropy principle~\cite{Ziebart2008MaximumEI} based on the value $V^{\pi}$ as ${P(V^{\pi}) = Z^{-1} \mbox{exp}(-\beta V^{\pi})}$,
    where the partition function is defined by ${Z=\sum_{\pi \in \Pi}{\mbox{exp}(-\beta V^{\pi})}}$ and the temperature parameter $\beta$.
    The planner acts as an approximation method to the partition function similar to Markov chain Monte Carlo methods~\cite{rosbachDrivingStyleInverse2019}.
    We discard collisions in the calculation of the distribution to focus on the kinematically feasible, non-colliding set for which the features of the reward function provide a coherent ranking.
    We manually tune the temperature $\beta$ to balance the planner solution bias and uncertainty of the distribution in the reachable set.
    
    Each pixel in each layer image receives the maximal value of any policy passing through that pixel at any time. 
    The resulting function for each pixel's color value can be written as
    \begin{gather}
    V(p_x, p_y, l) = \max_{\pi \in \Pi, t\in T_l} P_\pi(p_x, p_y, t) \\
    p_x \in \{1,\ldots,\textrm{width}\} ,~ p_y \in \{1,\ldots,\textrm{height}\}.
    \label{eq:optimization}
    \end{gather}

    We calculate pixel coordinates from Cartesian coordinates of the policy every 0.2 seconds, which provides start and end points for a Bresenham~\cite{bresenham1965algorithm} line drawing algorithm connecting the pixels.
    The rendering algorithm is explained in Algo.~2. 

    \begin{figure}
    \vspace{1.8mm}
    \centering
    \includegraphics[align=t,width=0.49\textwidth]{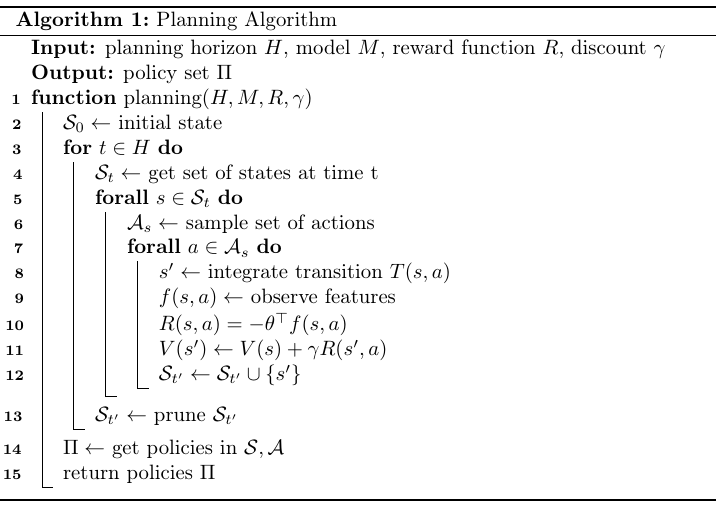}
    ~
    \includegraphics[align=t,width=0.49\textwidth]{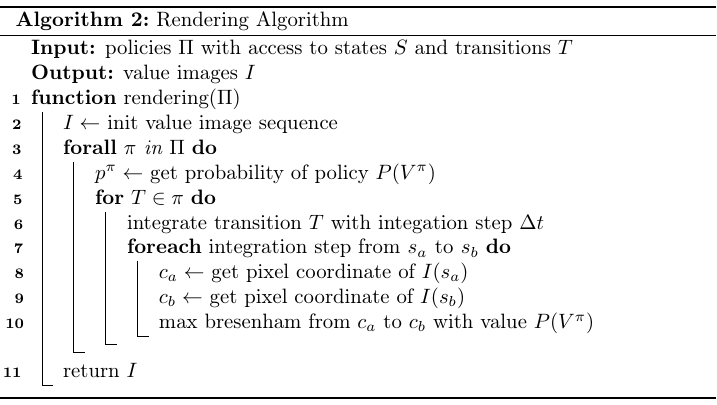}

    \end{figure}

\subsection{Image-to-Image Prediction}

    We propose a conditional Generative Adversarial Network (GAN) training methodology to perform image-to-image prediction.
    The deep neural network architecture utilized is based on an Encoder-Decoder U-Net architecture as in Pix2Pix~\cite{isolaImagetoImageTranslationConditional2018}.
    The network is modified to take input images of size $512 \times 512 \times 4$ and produce images of size $512\times 512 \times 12$ in contrast to the original architecture using a spatial resolution of $256\times 256$.
    Target images are scaled to the interval $[0,1]$ with a cut-off at $0.1$ for the lowest possible reachable values.
    The generator is trained using hard sigmoid activation functions to assign values of zero to pixels out of the time-dependent non-colliding reachable set and values of one to the best pixels of the set. 
   
    The capacity of the generator is increased by using an additional down- and upsampling layer, each having $512$ filters resulting in a total of $67$~M parameters. The discriminator has an additional downsampling layer with $512$ filters with a total of $7$~M parameters.
    We resize the $512\times 512$ square output images again to the target aspect ratio, resulting in a resolution of one megapixel.
    The neural network must adapt to various scaling factors due to the planning algorithm's situation-dependent choices regarding the meters-to-pixel scale and image aspect ratio.
    
\subsection{Planning with the PSVN}

    When using the PSVN the infrastructure and objects are provided in bird's eye view representation as depicted in Fig.~\ref{fig:title}. 
    The PSVN directly provides pixel state values $V(p_x,p_y,l)$ in a sequence of images encoding values for planning, replacing explicit object predictions that are usually required to plan in interactive environments.
    Recall that pixel values are represented in the scaled interval $[0,1]$, where a value of $0$ represents unreachable or colliding pixel states and $1$ pixels being part of the optimal policy having received the $\max_{\pi \in \Pi, t\in T_l} P_\pi(p_x, p_y, t)$ pixel update.
    This work uses the pixel values to perform reward shaping for our planning algorithm listed in Algo.~1 to improve the optimal policy selection~\cite{ng1999policy}.
    Therefore supplementing the missing behavioral information in features $f(s,a)$ of the reward function that only depend on $s$ and action $a$ with the PSVN values encoding the preference of following a desired behavior in a situation.
    The new reward function of the planner is given by $R'= R + F$, where the reward shaping function $F$ is obtained by reading and integrating pixel values encountered on the continuous transition from state $s$ to $s'$ using $log(V(p_x,p_y,l))$.
    The continuous transitions are integrated using a vehicle model, and at every intermediate step, the Cartesian state coordinates are converted into pixel coordinates.
    The new reward function $R'$ combines prior knowledge such as vehicle kinematics and distances to infrastructure in $R$ with situation-dependent abstract behavioral information given by the reward shaping function $F$ such as how the ego vehicle has to behave with respect to other objects. 
    
\section{Experiments}
\label{sec:result}
    We show experiments that exhibit the combined prediction and planning capabilities of the proposed network in interactive driving environments.
    Our experiments are divided into a qualitative analysis of the predicted image sequences and a quantitative evaluation of the policy selection.
    The quantitative evaluation contrasts the performance of a traditional architecture separating prediction and planning with the proposed approach combining PSVN with planning.
    The training dataset contains 24 hours of real-world driving data from 35 different drivers on a predetermined route.
    The driving data includes mostly urban situations with a short rural road section.
    Following the route requires multiple lane changes, left and right turns at traffic lights, and roundabouts covering each exit possibility.
    The datasets are filtered discarding situations having less than one other vehicle involved.
    In the following experiments, we perform a simplification of the implementation by using an object prediction module as described in the traditional sequential chain to render the training dataset targets.
    This allows us to disregard perception and tracking uncertainties at this stage of this work.
    Furthermore, this simplifies the quantitative evaluation allowing a direct comparison of the separated prediction and planning architecture with the proposed approach combining PSVN with planning.
    We augment the data by shifting the ego vehicle forward, closer to preceding vehicles, increasing the ego state velocity, and randomizing the position and orientation relative to the centerline.
    All other input information remains unmodified.
    The training dataset is rebalanced based on the aspect ratio of the target images, which provide a proxy to different situations, e.g., squared images in roundabouts and rectangular in straights.
    Inputs for tests are generated in a simulation in separate geographical locations compared to the training dataset covering similar situations.
    The test dataset contains a roundabout and multi-lane road segments. 
    Lane changes are required to follow the route and adhere to user inputs such as preference for adjacent lanes.
    Setting the preference for neighboring lanes in the presence of other moving vehicles allows us to generate difficult scenarios with conflicting objectives between route navigation and object interaction.


\subsection{Qualitative Examples}
\label{sec:qualitative}
    
\begin{figure*}
\vspace{1.8mm}
\centering
\begin{subfigure}[b]{.23\textwidth}
\begin{tikzpicture}
\node[above right, inner sep=0] (image) at (0,0) {
\includegraphics[width=\textwidth]{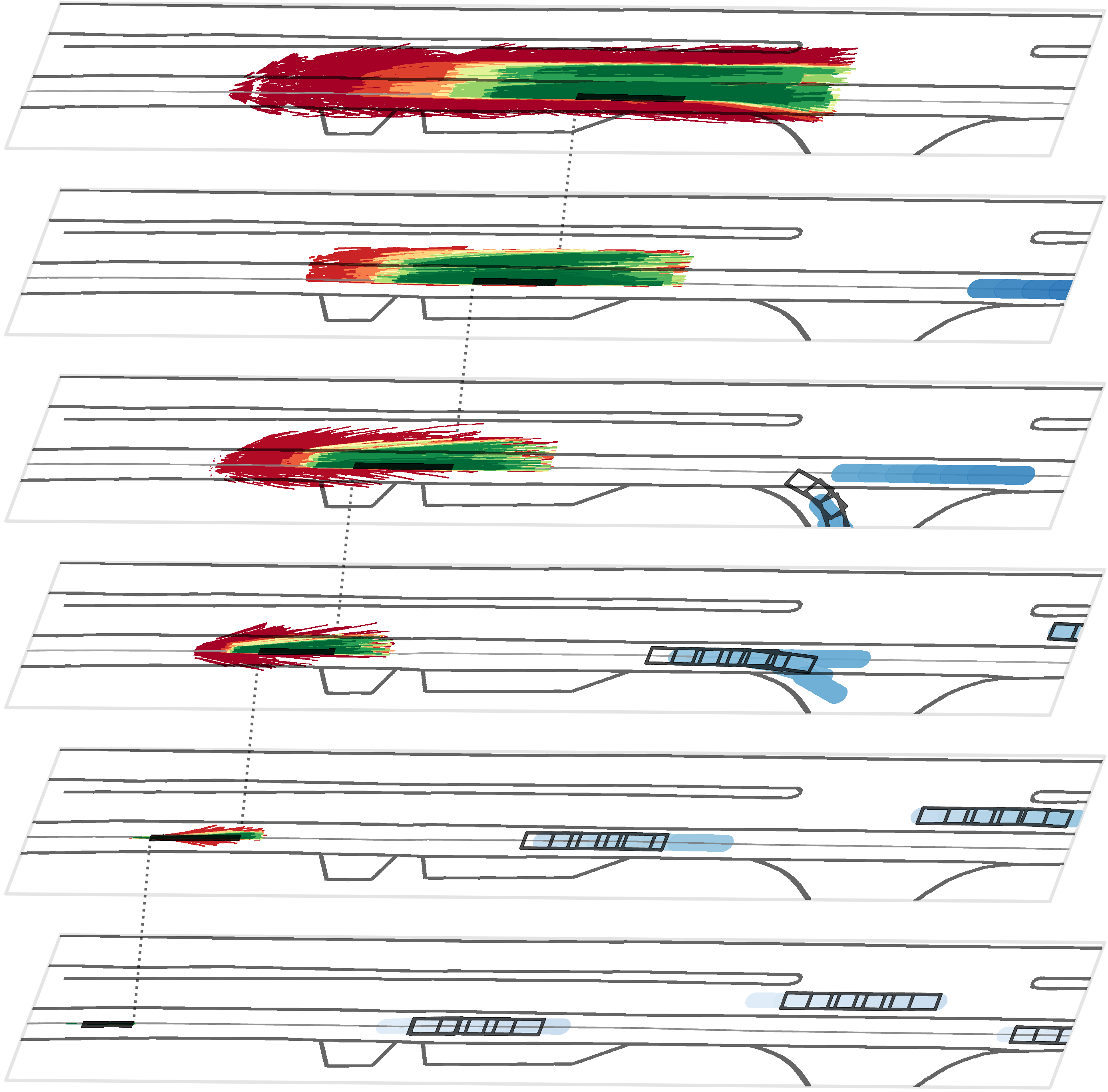}};
 \draw[-latex, line width=0.5pt,opacity=0.5] (-0.1,0.1) -- (-0.1,3.4);
 \node[above right, rotate=90, opacity=0.5] at (0,0) {\tiny 6 time layers in 1.1 sec. intervals};
\end{tikzpicture}
\caption{Straight Target}
\label{fig:straight_target}
\end{subfigure}
\hspace{3.8mm}
\begin{subfigure}[b]{0.23\textwidth}
\includegraphics[width=\textwidth]{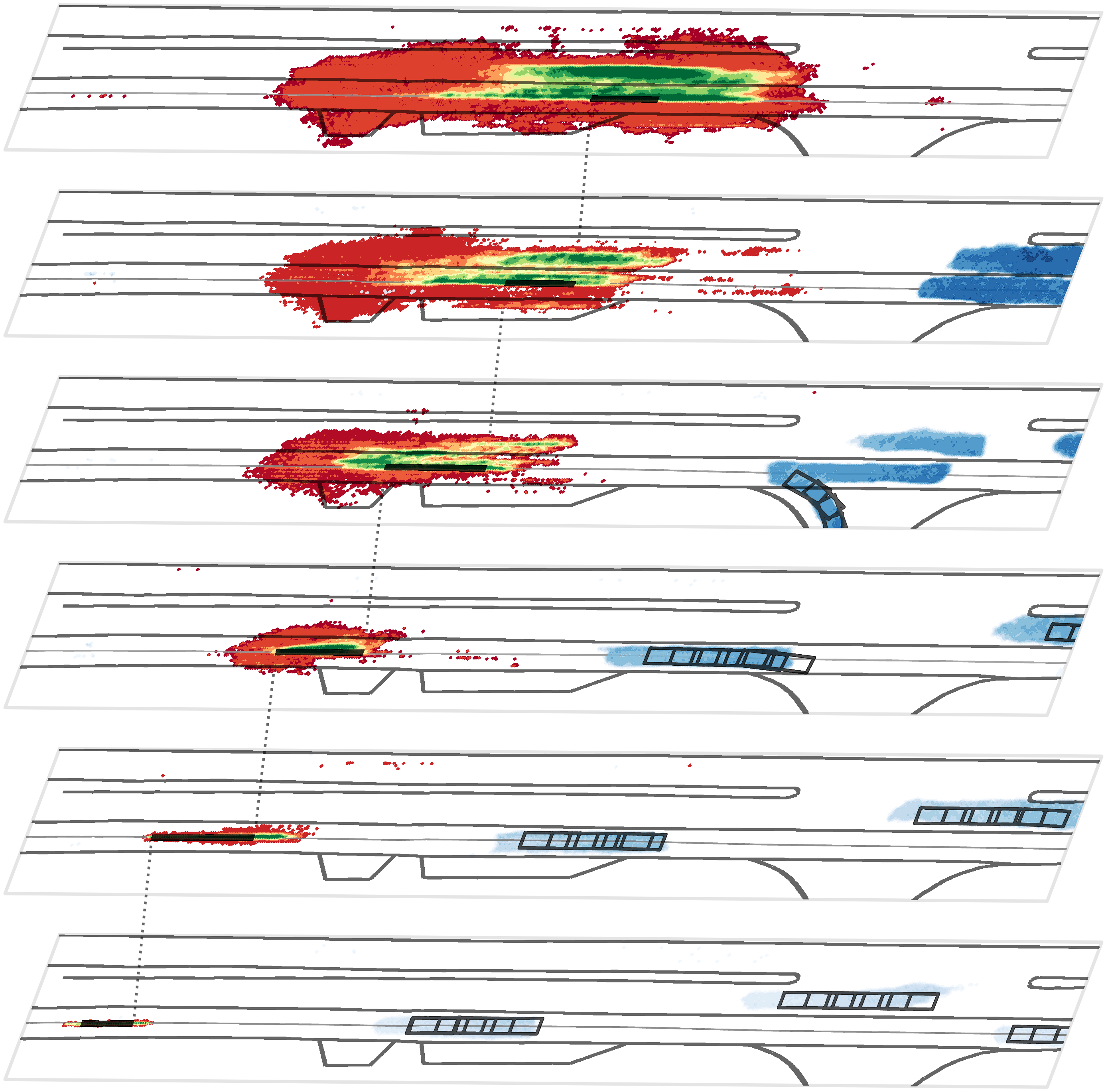}
\caption{Straight Prediction}
\label{fig:straight_pred}
\end{subfigure}
\begin{subfigure}[b]{0.24\textwidth}
\includegraphics[width=\textwidth]{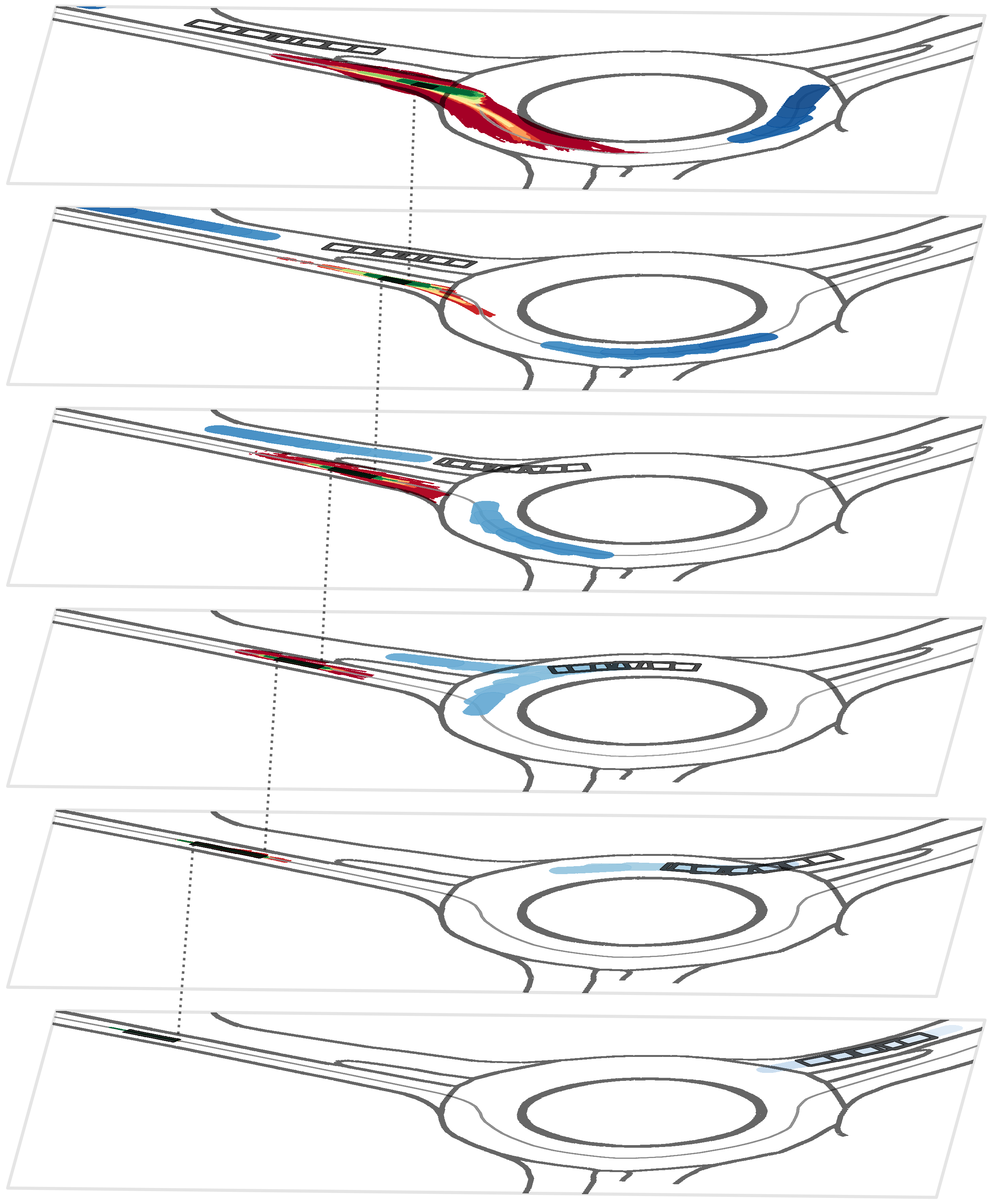} 
\caption{Roundabout Target}
\label{fig:kreisel_target}
\end{subfigure}
\begin{subfigure}[b]{0.24\textwidth}
\includegraphics[width=\textwidth]{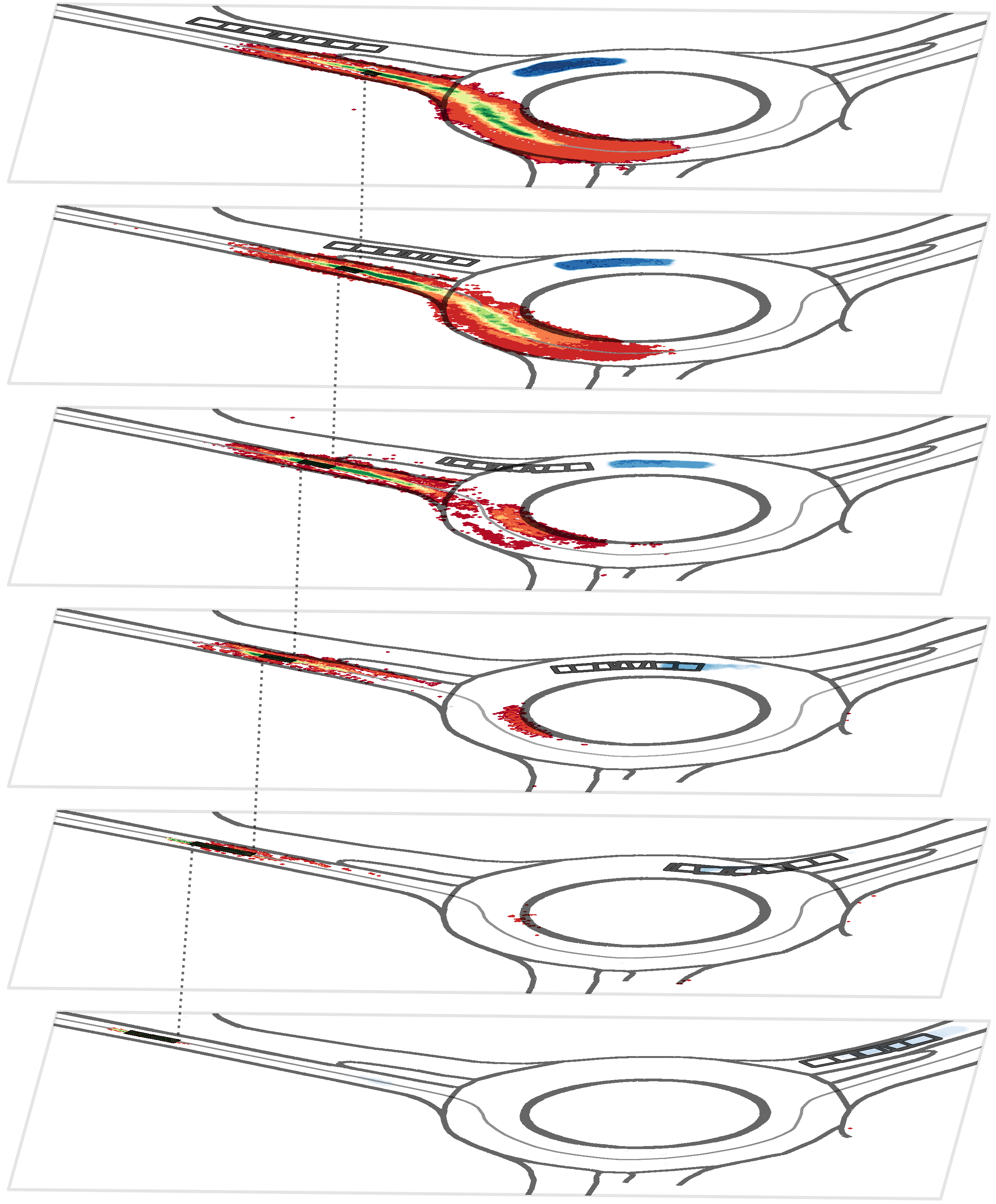} 
\caption{Roundabout Prediction}
\label{fig:kreisel_pred}
\end{subfigure}
\begin{subfigure}[b]{.24\textwidth}
\includegraphics[width=\textwidth]{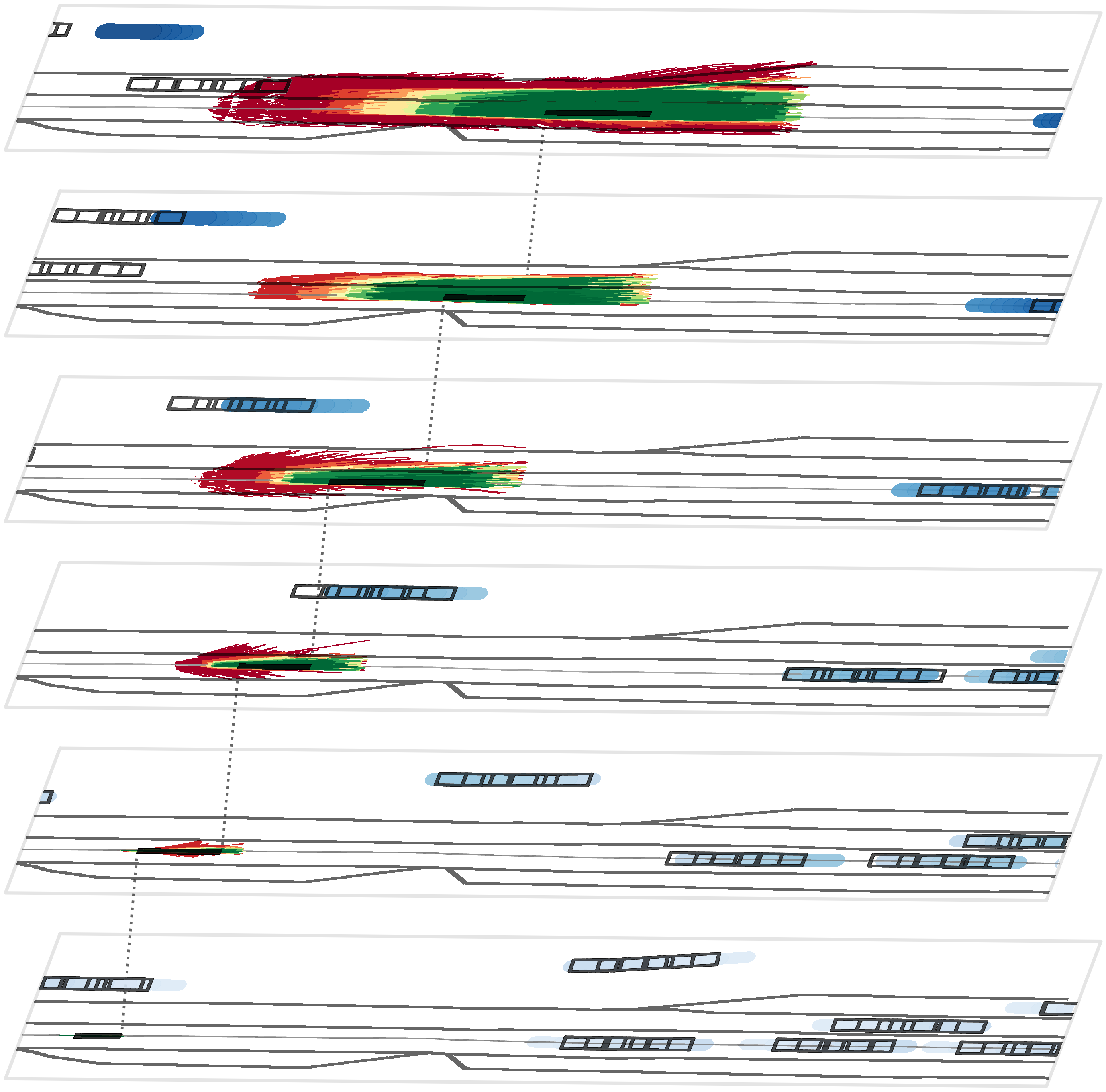}
\caption{Straight Target}
\label{fig:many_target}
\end{subfigure}
\begin{subfigure}[b]{.24\textwidth}
\includegraphics[width=\textwidth]{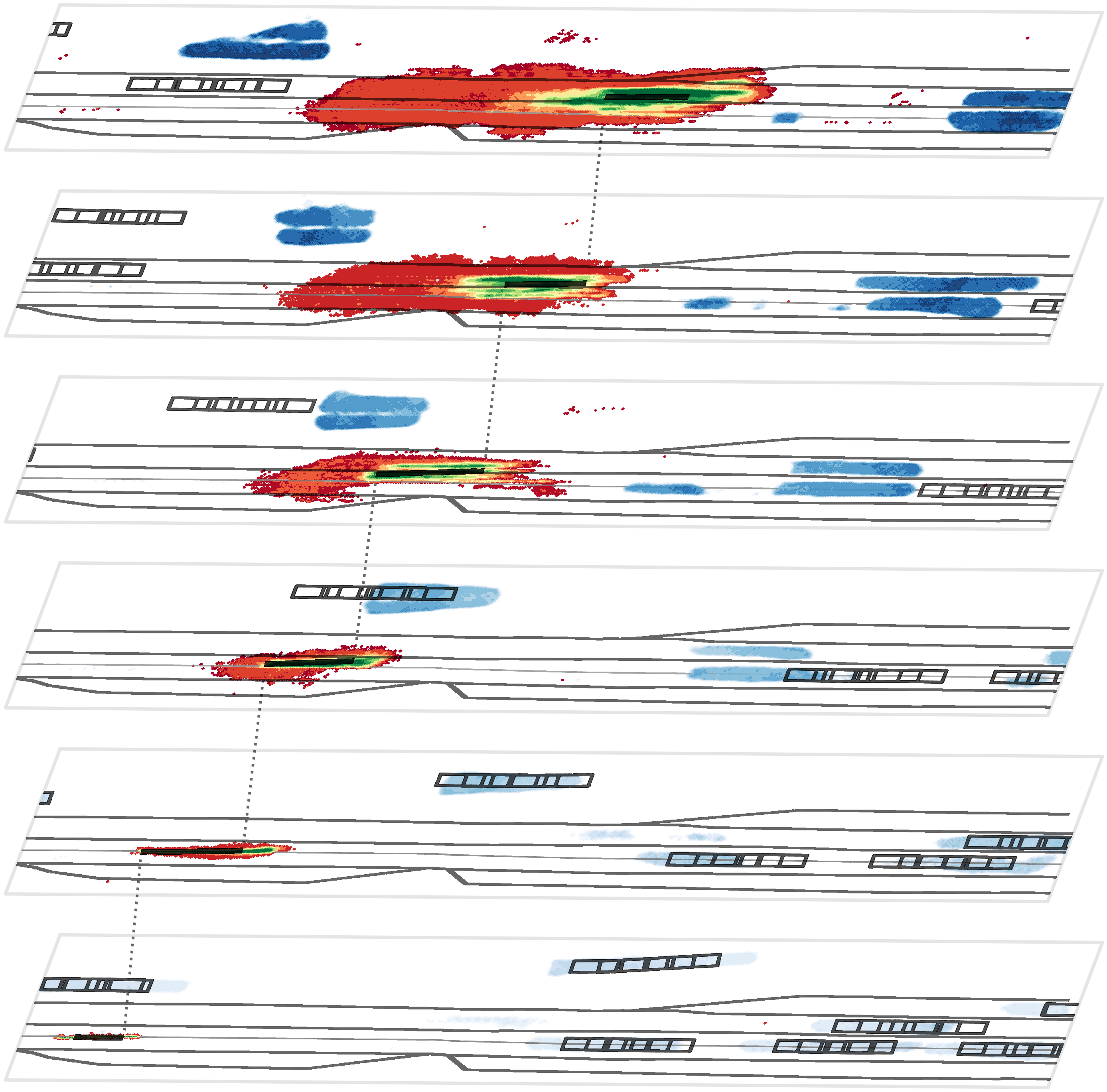} 
\caption{Straight Prediction}
\label{fig:many_pred}
\end{subfigure}
\label{fig:qual1}
\hfill
\begin{subfigure}[b]{.24\textwidth}
\includegraphics[width=\textwidth]{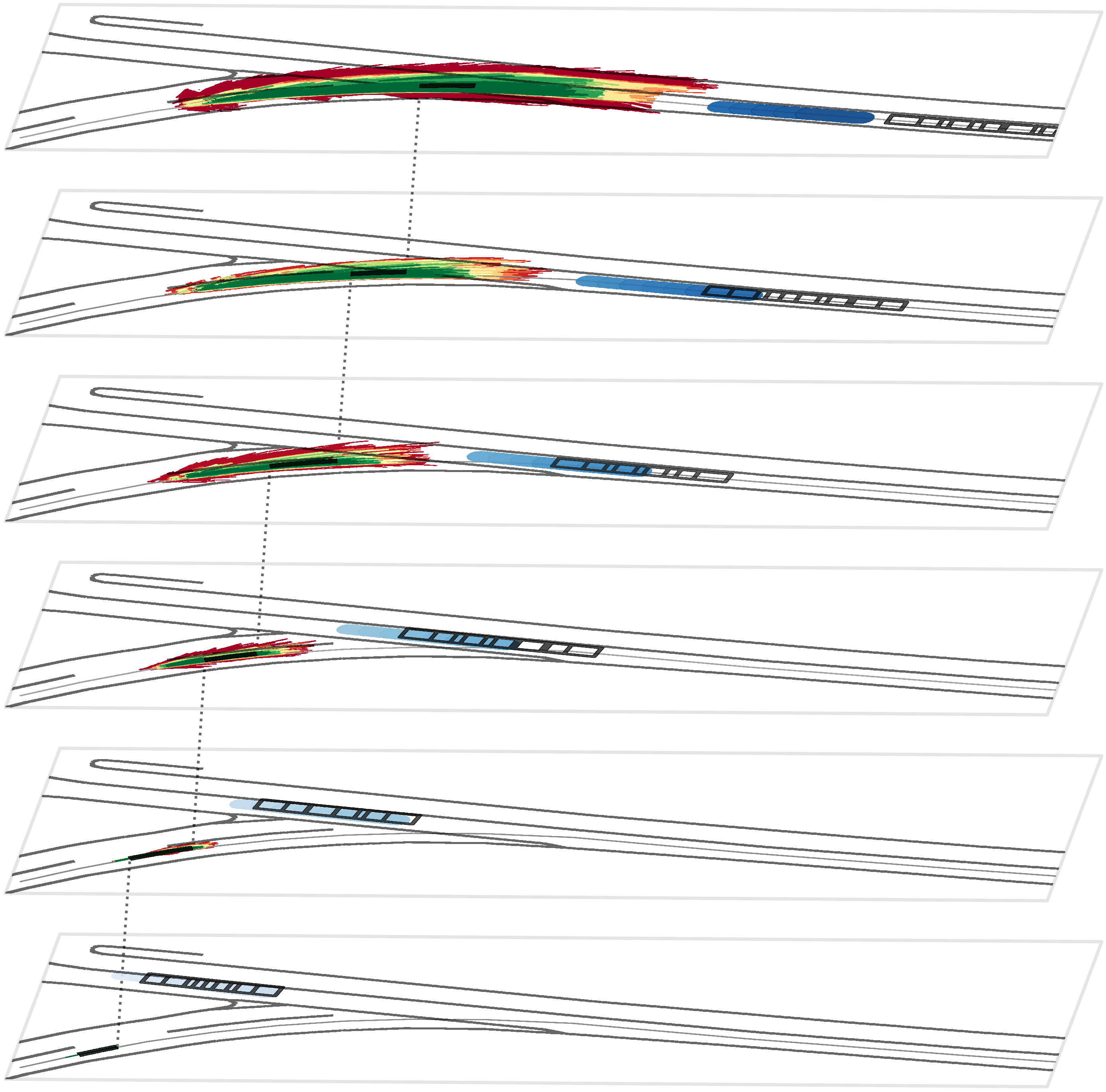}
\caption{Merge Target}\label{fig:merge_target}
\end{subfigure}
\begin{subfigure}[b]{.24\textwidth}
\includegraphics[width=\textwidth]{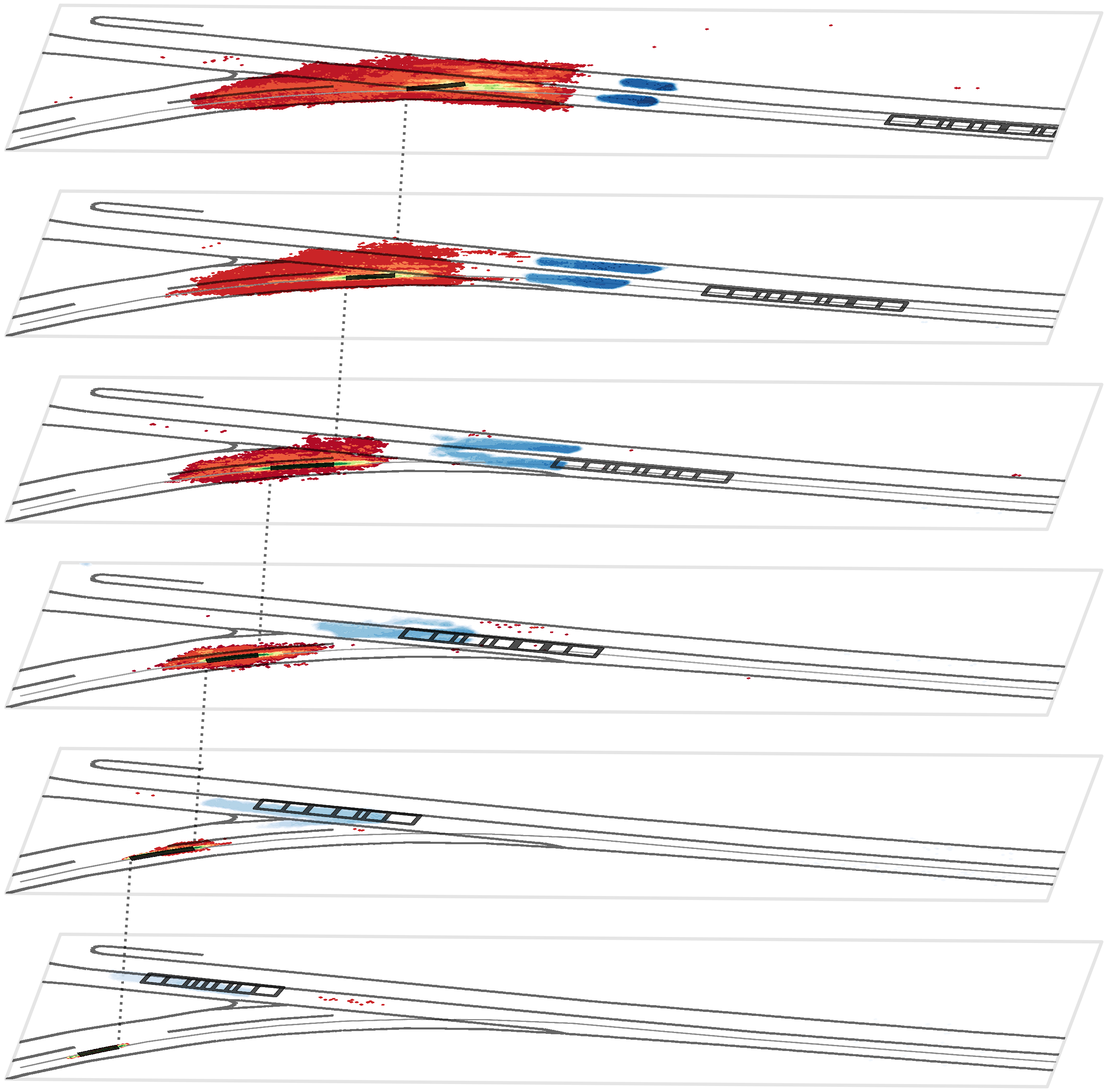}
\caption{Merge Prediction}\label{fig:merge_pred}
\end{subfigure}

\caption{
    Displays targets \emph{(a,c,e,g)} and predictions \emph{(b,d,f,h)} for straights, roundabout and merge situations.
    Motion predictions of objects are displayed in blue color.
    State values are depicted in red, yellow, and green colors.
    Multiple overlays are displayed in grey (boundaries, target lane, ground-truth objects).
    The trajectories passing through time layers are drawn in black. The trajectories in \emph{(a,c,e,g)} are selected based on the predicted state value images and in \emph{(b,d,f,h)} are based on the trajectory values.}
\label{fig:qual2}
\end{figure*}

    First, we show qualitative examples of the combined prediction of object motion and ego vehicle behavior in interactive urban traffic environments.
    The images contrast the model's prediction with rendered target images.
    Both prediction and target are depicted including overlays of ground-truth object positions (black rectangles), infrastructure, and the selected trajectory in black passing through all time layers.
    We show a few selected situations with parallel and merging lanes, as well as a roundabout, which require particular emphasis on object motion prediction and timing of the interactive behavior.
    Figs.~\ref{fig:straight_target} and \ref{fig:straight_pred} display an often encountered situation with two parallel lanes, where the ego vehicle has to follow a preceding vehicle.
    The preceding vehicle might take the exit or continue heading in the same direction as the ego vehicle.
    Both prediction and target provide multi-modal behavior in state values and object motion and both choose to follow the vehicle in the current lane instead of an unnecessary lane change.
    Figs.~\ref{fig:many_target} and \ref{fig:many_pred} depict a similar situation with both lanes occupied by preceding vehicles.
    The network correctly distinguishes objects based on their driving direction and predicts the oncoming traffic.
    However, the timing of the explicit motion predictions do not match the ground truth; specifically, the network estimates objects to move slower.
    The optimal policy selected sorely on predicted values performs a lane change, which is feasible in this situation to gain more progress along the route.
    Figs.~\ref{fig:merge_target} and \ref{fig:merge_pred} depict a merging situation, where the ego vehicle has to judge the velocity and direction of the moving vehicle in order to determine whether to move behind or in front of the vehicle.
    Our network is able to generate a similar motion prediction as the target until 3.3 sec., as depicted in the third time-dependent image.
    Thereafter, the proposed network generates two hypotheses for the motion prediction and positions itself behind.
    Figs.~\ref{fig:kreisel_target} and \ref{fig:kreisel_pred} demonstrate a roundabout with the object taking the first exit.
    The predicted state values focus around the centerline of the roundabout, displaying that arbitrary geometries can potentially be predicted.
    The solution candidate of the prediction shows preference for not entering the roundabout until more information is revealed of the objects' intention.

\subsection{Quantitative Evaluation}
\label{sec:quantitative}

    In our quantitative evaluation, we contrast the policy selection of the planner using the training setup with object predictions available against the inference setup only using the PSVN as input.
    After sampling a policy set for each situation while neglecting object related features, we select two policies by computing the $\max_{\pi \in \Pi} V^{\pi}$.
    The first policy is selected by using the reward function $R'= R$ that includes explicit object prediction features (the traditional chain).
    The second policy is selected by solely using the reward shaping function $R'= F$ (the output of the PSVN).
    We purposefully neglect all state and action rewards $R(s,a)$ of $R'$ to analyze the encoding capabilities of the inferred pixel state values and focus on object interaction.
    Metrics that compare against ground truth trajectories are problematic for evaluation of interactive behavior as in most situations multiple behaviors are valid.
    Therefore we propose to use a different evaluation metric that is termed object time gap (OTG). 
    It is defined for a policy $\pi$ as
    
    \begin{gather*}
    \mathrm{OTG}(\pi) = \min_{\substack{obj \in O\\  t_{obj} \in [0,H]\\  t_{\pi} \in [0,H]}} 
    \left| t_\pi - t_{obj} \right|  s.t.
    \left \|
    \begin{pmatrix}
    x_{t_\pi} - x_{t_{obj}}\\ 
    y_{t_\pi} - y_{t_{obj}}
    \end{pmatrix} 
    \right \|^2 < 2,
    \label{eq:otg}
    \end{gather*}
    
    where $O$ is the set of all objects in that scenario, $x$ and $y$ are the Cartesian coordinates of an object or the ego vehicle at that time $t$.
    An object time gap of 0 sec.\ represents the ego vehicle and an object reaching the same place at the same time (collision). 
    A 3 sec.\ time gap means that one object and the ego vehicle reach the same place with a 3 sec.\ time difference. 
    If no path overlap exists, the object time gap results in a value of $\infty$.
    Fig.~\ref{fig:confusion:otg} shows in the upper right corner that in most scenarios both the trajectory selected by the planner reward function and the trajectory selected by the predicted values never overlap with objects. 
    It can be seen in our supplementary video that there are always vehicles present in the situations.
    This means that the selected policies do not display reckless behaviors such as driving into vehicles on parallel lanes.
    Many scenarios fall into the diagonal where both selected trajectories display correlating OTG. 
    Another accumulation can be seen on the first row, where the reward function based selection correctly avoids objects but the predicted value map produces more aggressive behavior.
    The second confusion matrix (see Fig.~\ref{fig:confusion:dist}) displays the travel distance of a trajectory relative to the initial position. 
    Values are accumulated slightly below the diagonal.
    It visualizes that in many scenarios the prediction based selection results in a further progressing trajectory in comparison to the reward function based selection.
    This shows that trajectories selected by our prediction value successfully leave large enough object time gaps or completely avoid collisions while not trivially standing still.

\begin{figure}
\centering
\begin{subfigure}[b]{.49\textwidth}
\pgfkeys{/pgf/number format/.cd,1000 sep={}} 
\begin{tikzpicture}
    \begin{axis}[
     width=7cm,
     height=5cm,
    font=\small,
            colormap={bluewhite}{color=(white) rgb255=(90,96,191)},
            xlabel=selected by PSVN,
            xlabel style={yshift=-2pt},
            ylabel=selected by reward function,
            ylabel style={yshift=2pt},
            xticklabels={$<\!1$, $<\!2$, $<\!3$, $<\!4$, $<\!5$, $<\!6$, $<\!7$, $\infty$},
            xtick={0,...,7}, 
            xtick style={draw=none},
            yticklabels={$\infty$, $<\!7$, $<\!6$, $<\!5$, $<\!4$, $<\!3$, $<\!2$, $<\!1$},
            ytick={0,...,7}, 
            ytick style={draw=none},
            enlargelimits=false,
            nodes near coords={\pgfmathprintnumber\pgfplotspointmeta},
            nodes near coords style={
                yshift=-7pt
            },
        ]
        \addplot[
            matrix plot,
            mesh/cols=8, 
            point meta=explicit,draw=gray,
            point meta max=300,
        ] table [meta=C] {
            x y C
            0 0 58
            1 0 59
            2 0 205
            3 0 160
            4 0 149
            5 0 132
            6 0 31
            7 0 2473
            0 1 0
            1 1 0
            2 1 1
            3 1 1
            4 1 2
            5 1 0
            6 1 0
            7 1 8
            0 2 0
            1 2 1
            2 2 1
            3 2 4
            4 2 13
            5 2 19
            6 2 1
            7 2 47
            0 3 4
            1 3 6
            2 3 9
            3 3 6
            4 3 33
            5 3 15
            6 3 2
            7 3 33
            0 4 17
            1 4 23
            2 4 45
            3 4 51
            4 4 47
            5 4 10
            6 4 2
            7 4 41
            0 5 76
            1 5 166
            2 5 175
            3 5 32
            4 5 10
            5 5 7
            6 5 1
            7 5 35
            0 6 65
            1 6 112
            2 6 68
            3 6 17
            4 6 5
            5 6 0
            6 6 0
            7 6 21
            0 7 100
            1 7 26
            2 7 45
            3 7 16
            4 7 6
            5 7 5
            6 7 0
            7 7 35
        }; 
    \end{axis}
\end{tikzpicture}
\caption{Object Time Gap [s]}\label{fig:confusion:otg}
\end{subfigure}
\hfill
\begin{subfigure}[b]{.49\textwidth}
\begin{tikzpicture}
    \begin{axis}[
     width=7cm,
     height=5cm,
    font=\small,
            colormap={bluewhite}{color=(white) rgb255=(90,96,191)},
            xlabel=selected by PSVN,
            xlabel style={yshift=-2pt},
            ylabel=selected by reward function,
            ylabel style={yshift=2pt},
            xticklabels={$<\!10$, $<\!20$, $<\!30$, $<\!40$, $<\!50$, $<\!60$, $<\!70$, $>\!70$},
            xtick={0,...,7}, 
            xtick style={draw=none},
            yticklabels={$>\!70$, $<\!70$, $<\!60$, $<\!50$, $<\!40$, $<\!30$, $<\!20$, $<\!10$},
            ytick={0,...,7}, 
            ytick style={draw=none},
            enlargelimits=false,
            nodes near coords={\pgfmathprintnumber\pgfplotspointmeta},
            nodes near coords style={
                yshift=-7pt
            },
        ]
        \addplot[
            matrix plot,
            mesh/cols=8, 
            point meta=explicit,draw=gray,
            point meta max=300,
        ] table [meta=C] {
                    x y C
                    0 0 0
                    1 0 0
                    2 0 35
                    3 0 67
                    4 0 76
                    5 0 71
                    6 0 125
                    7 0 820
                    0 1 0
                    1 1 1
                    2 1 6
                    3 1 4
                    4 1 7
                    5 1 3
                    6 1 10
                    7 1 45
                    0 2 0
                    1 2 0
                    2 2 6
                    3 2 8
                    4 2 6
                    5 2 12
                    6 2 8
                    7 2 25
                    0 3 1
                    1 3 3
                    2 3 14
                    3 3 20
                    4 3 32
                    5 3 21
                    6 3 46
                    7 3 272
                    0 4 4
                    1 4 10
                    2 4 91
                    3 4 92
                    4 4 82
                    5 4 106
                    6 4 161
                    7 4 673
                    0 5 12
                    1 5 31
                    2 5 54
                    3 5 56
                    4 5 66
                    5 5 70
                    6 5 74
                    7 5 117
                    0 6 56
                    1 6 101
                    2 6 142
                    3 6 175
                    4 6 168
                    5 6 134
                    6 6 95
                    7 6 79
                    0 7 163
                    1 7 79
                    2 7 43
                    3 7 36
                    4 7 13
                    5 7 3
                    6 7 2
                    7 7 0
        }; 
    \end{axis}
\end{tikzpicture}
\caption{Progress of Trajectory [m]}\label{fig:confusion:dist}
\end{subfigure}
\caption{
    Confusion matrices contrasting optimal policy of planner using the reward function and PSVN for 4732 test scenarios that each contain objects. 
}
\label{fig:confusion}
\end{figure}

\section{Limitations and Future Work}
\label{sec:limitations}

As of now, this work assumes access to an a-priori defined reward function with features encoding information about ego motion, infrastructure, and dynamics of the environment.
The reward function for planning is a linear weighing of the relevance of these features.
We aim to increase the quality of rendered target images by using situation-dependent reward function parameters~\cite{rosbachDrivingStyleEncoder2020,rosbach2020planning}.
We see a lot of potential in substituting the segmentation focused network architecture with a recurrent architecture that is more capable of capturing dynamics.
We have shown that our training methodology is able to generate high-resolution situation-dependent state values and motion predictions for a diverse set of environments.
We see a lot of potential when using our methodology to pretrain a network to generate an encoding of the situation and fine-tune the network for other tasks such as predicting actions or reward functions.
We proposed using PSVN for reward shaping, however aim to incorporate predicted pixel state values as admissible heuristic to accelerate search and reduce the GPU memory requirements for the set of states and actions when deploying in vehicle.

\section{Conclusion}
\label{sec:conclusion}

This work proposes a methodology for combined pixel state value and object motion prediction.
A conditional GAN is trained to predict both a sequence of value and object motion images given a bird's eye view of the environment.
The proposed planning and rendering algorithms make the generation of one megapixel-sized pixel state value image sequences tractable.
This allows training the generative model with large image pair datasets.
Our training algorithm uses consumer grade hardware and does not require interaction with the environment, which allows us to utilize real-world driving recordings.
The results demonstrate confident multi-modal pixel state value and object predictions in situations requiring moderate timing capabilities amidst a large set of moving objects and a promising outlook for complex interactive situations such as roundabouts and merging situations.

\addtolength{\textheight}{-7cm}

\section*{Acknowledgments}
This research was supported by Federal Ministry for Economic Affairs and Climate Action in the national research project RUMBA under grant number 19A20007L.
\newpage

\bibliographystyle{IEEEtran}
\bibliography{main}

\end{document}